\documentclass[12pt,twocolumn,letterpaper]{article}
\usepackage{cvpr}
\usepackage{graphicx}
\usepackage{hyperref}
\usepackage{tabularx}
\usepackage{times}
\usepackage{epsfig}
\usepackage{graphicx}
\usepackage{amsmath}
\usepackage{amssymb}
\usepackage[font=small,labelfont=bf]{caption}
\usepackage{cite}
\usepackage[flushleft]{threeparttable}

\newcolumntype{s}{>{\hsize=1.25\hsize}X}

\usepackage{titlesec}
\titleformat{\section}{\normalfont\Large\bfseries}{\thesection}{1em}{}
\titleformat{\subsection}{\normalfont\large\bfseries}{\thesubsection}{1em}{}
\titleformat{\subsubsection}{\normalfont\normalsize\bfseries}{\thesubsubsection}{1em}{}

\cvprfinalcopy 

\def\cvprPaperID{****} 

\title{\vspace{-0.5cm}Addressing Data Misalignment in Image-LiDAR Fusion 
\linebreak on Point Cloud Segmentation\vspace{-0.5cm}}

\author{
    \begin{tabular}{cc}
        Wei JongYang & Guan Cheng Lee\\
        {\tt\small WJYang@ncut.edu.tw} & {\tt\small nm6111027@gs.ncku.edu.tw}
    \end{tabular}
}

\begin{document}

\maketitle

\begin{abstract}
    With the advent of advanced multi-sensor fusion models, there has been a notable enhancement 
    in the performance of perception tasks within in terms of autonomous driving.
    Despite these advancements, the challenges persist, particularly in the fusion of data 
    from cameras and LiDAR sensors. A critial concern is the accurate alignment of data from these 
    disparate sensors. Our observations indicate that the projected positions of LiDAR points often misalign
    on the corresponding image.
    Furthermore, fusion models appear to struggle in accurately segmenting these misaligned points.
    In this paper, we would like to address this problem carefully,
    with a specific focus on the nuScenes dataset and the SOTA of fusion models 2DPASS,
    and providing the possible solutions or potential improvements.
\end{abstract}

\section{Introduction}
Utilizing a single modality model is a prevalent approach when addressing perceptual challenges in deep learning.
However, for systems that demand high accuracy and robustness, such as autonomous driving, this approach may not suffice.
Without a doubt, autonomous driving perception has emerged as one of the most scrutinized areas of research in recent times.
Given the paramount importance of safety and reliability in this domain, the fusion of multi-sensor data has become a must.

\begin{figure}[!t]
    \includegraphics[width=\linewidth]{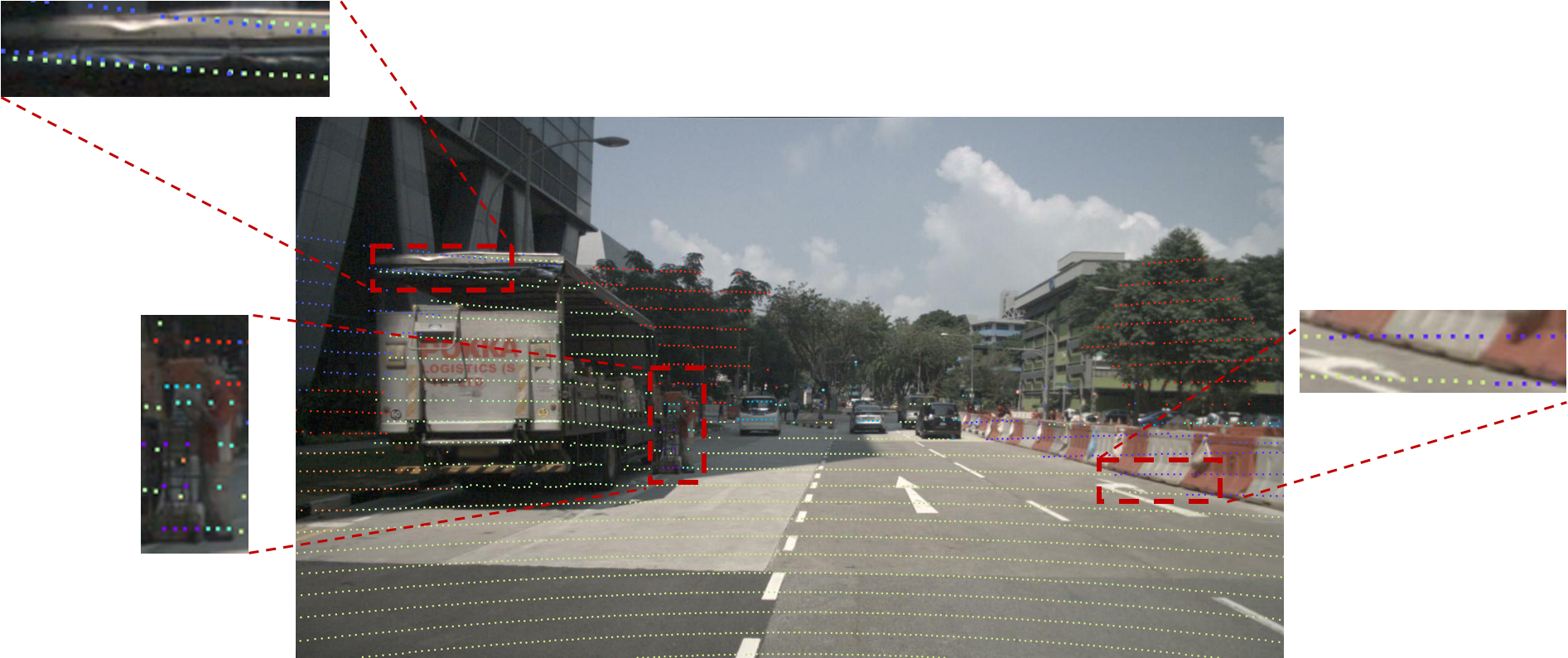}
    \centering
    \caption{NuScenes projection misalign example.
    The different color of these points represent the category they belong to.
    In the left side of this image, the points of the building behind the truck are overlapped with the truck,
    and the points of barriers on the right are projected onto the road.
    }
    \label{fig:nuScenes_mismatch_projection}
\end{figure}

The task of fusing multiple modalities presents significant challenges, the priorof which is the alignment of data from disparate sensors.
Coordinate transformation serves as a standard preprocessing step to align data between modalities, for example, transforming LiDAR points to align with the camera's image plane.
However, the synchronization between the LiDAR and camera is not precise, LiDAR points might projected onto wrong pixels.
The existing researches on this topic have not addressed this issue adequately, even the best models in the open leaderboard.
nuScenes \cite{caesar2020nuscenes} is the most popular dataset for autonomous driving perception tasks, 
it provide multiple modalities data, including LiDAR, camera, radar, and GPS, and rich annotations, 3D bounding boxes, and point-level segmentation labels, etc.
However, the projected positions of LiDAR points often misalign on the corresponding image since varying reasons,
the example is shown in Figure \ref{fig:nuScenes_mismatch_projection}, the official API of nuScenes also provide the same projection result.
We also found that SemanticKITTI\cite{behley2019semantickitti} show the same problem, but it is not as serious as nuScenes.

\cite{yan20222dpass, zhuang2021perception} are the top two opensourced models in the nuScenes LiDAR segmentation challenge in terms of multi-modality,
2DPASS\cite{yan20222dpass} is one of the best model in the leaderboard, achieve 80.8 mIoU on the nuScenes test set, and 72.9 mIoU on SemanticKITTI test set.
Nevertheless, all these three models do not address the misalignment problem in model design and dataset perspective,
and we found that 2DPASS might fail to segment the misaligned points more often.

In the following sections, we will address how the LiDAR is projected onto the image, the reason why the misalignment problem occurs,
and discuss the possible solutions or potential improvements.

\section{Problem Definition}
In this section, we will first show that how LiDAR points are projected onto the image,
and then discuss the possible reasons why the misalignment problem occurs, finally, we will discuss the impact of the misalignment problem.

\subsection*{Coordinate Transformation and Rasterization}
For a given LiDAR point cloud $P=\{p_i\}_{i=1}^{N}$ and a given image plane $I=\{{{pix}_i}\}_{i=1}^{M}$, where $N$ is the number of points in the point cloud scene, M is the number of pixels of the image.
Assuming Point Cloud and RGB images are captured at the exactly same time, the objective is to rasterize the 3D points $p_i$ onto ${pix}_i$ 
by using a series of matrices multiplying, the general camera calibration formula is given as:
\begin{equation}
    [u_i, v_i, 1]^T = \frac{1}{z_i} \times K \times T \times [x_i, y_i, z_i, 1]^T ,
\end{equation}
where $K \in \mathbb{R}^{3 \times 4}$ is the camera intrinsic matrix 
and $T \in \mathbb{R}^{4 \times 4}$ is the camera extrinsic matrix,
the $T$ transforms the world coordinates into camera coordinates, the K transforms the camera coordinates to pixel coordinates.

In SemanticKITTI\cite{behley2019semantickitti} dataset, $K$ and $T$ has been directly provided.
Because of the difference between the frequency of capturing data of LiDAR and camera in NuScenes \cite{caesar2020nuscenes}, we need to consider the timestamp of the extrinsic matrices, the $T$ is given as:
\begin{flalign}
\begin{aligned}
    T=T_{camera \gets ego_{t_c}} \times T_{ego_{t_c} \gets global} \\ \times T_{global \gets ego_{t_l}} \times T_{ego_{t_l} \gets LiDAR} ,
\end{aligned}
\end{flalign}
where the subscripts marks what the coordinates transform from and to, for instance,
$T_{camera \gets ego_{t_c}}$ is the transformation matrix to project the view of the vehicle position at timestamp $t_c$ to its camera view, and so on.

The above process is widely addressed in the other related works \cite{yan20222dpass}.

\subsection*{Why the Misalignment Problem Might Occur}
The problem is also known as Sensors Time Synchronization Problem, the sensors are not synchronized in time.
The misalignment problem might occur due to the following reasons:
\begin{enumerate}
    \item The different operating frequency of camera and LiDAR
    \item The trigger mechanism of camera, LiDAR, GPS and IMU
    \item The timestamp delay between camera, LiDAR, GPS and IMU
    \item The noise of intrinsic and extrinsic parameters
\end{enumerate}
these factors tangled together and cause the misalignment problem, note that these are not the only reasons, but the most important ones,
\cite{9966179} address the other reasons in detail.

Let's just take the first factor as an example, the LiDAR and camera are not synchronized in time,
in nuScenes, for example, the LiDAR works at 20Hz, and the camera works at 12Hz, so the 12 camera exposures are speard
evenly over the 20 LiDAR sweeps, assume the car is moving ar 40 km/h, the longest distance between two is 7.3 meters, this is not even considering other factors.

Unfortunately, these hardware limitations can not be solved in the near future, the possible way to reduce the impact of these limitations is via model design perspective.

\begin{figure}[t]
    \includegraphics[width=\linewidth]{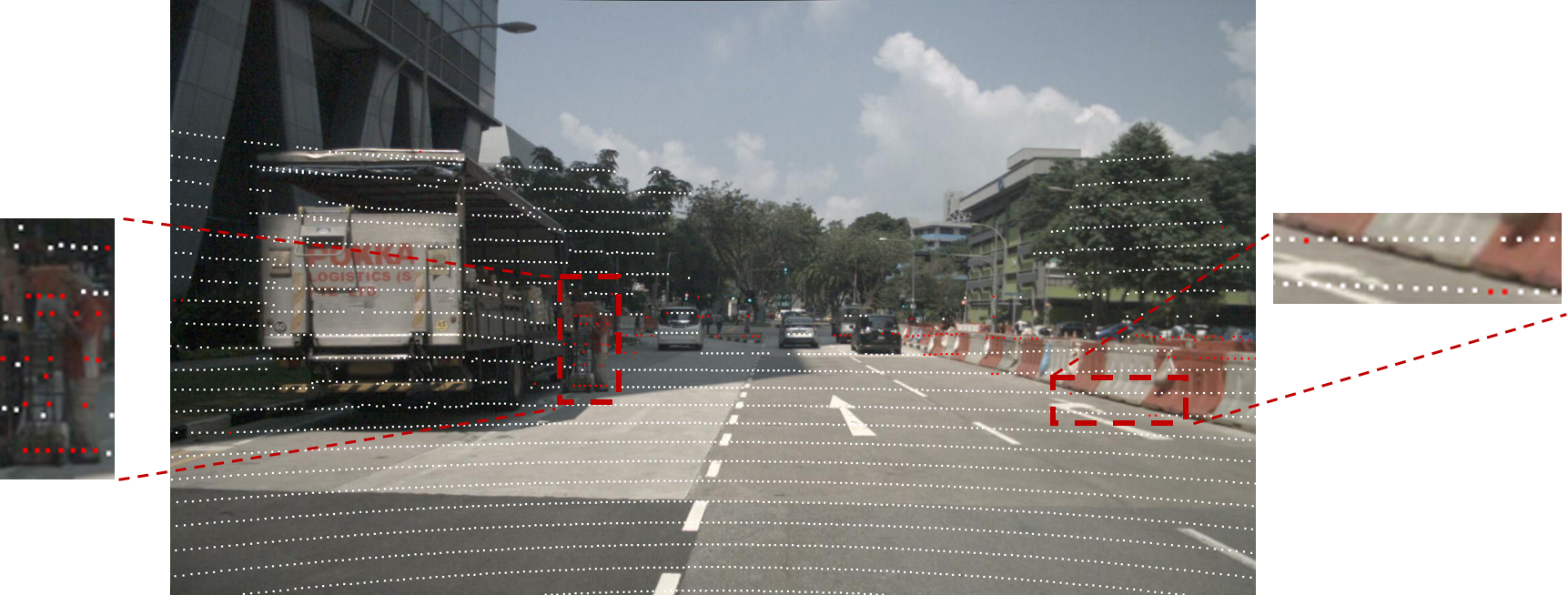}
    \centering
    \caption{The inference results of 2DPASS in nuScenes.
    }
    \label{fig:2DPASS_error}
\end{figure}

\subsection*{The Impact of Misalignment Problem to Deep Learning Models}

In \cite{zhuang2021perception}, they use the approach we show in section 2.1 to project the LiDAR points onto a image plane,
and concatenate with the corresponding RGB image, then feed the concatenated data into the RF fusion module. Which clearly would suffer from the misalignment problem.

The things are even clear in 2DPASS, each LiDAR point feature will concatenate with a corresponding pixel feature, unlike \cite{zhuang2021perception},
even though the image is not used during inference stage, the knowledge distillation information is inevitably to be noisy by the misaligned pixel features.

In Figure \ref{fig:2DPASS_error}, we show the inference results of 2DPASS in nuScenes, the red poins represent the wrong segmentation results.
Compare with misaligned points in Figure \ref{fig:nuScenes_mismatch_projection}, we can see that 2DPASS fails to segment on the misaligned points a lot,
but in some cases, the building behind the truck is segmented correctly, this is because the points are actually far from the truck, the 3D features lead the segmentation results.

\section{Possible Solutions and Potential Improvements}
Utilizing the right image and LiDAR data is the key to solve the misalignment problem,
unfortunately, it is not possible to know the specific root cause of every misaligned points.
A feasible way is to design a model that is adaptive to the misalignment points, select the right pixel features by itself.

\subsection*{Architecture of 2DPASS Overview}
\begin{figure}
    \includegraphics[width=\linewidth]{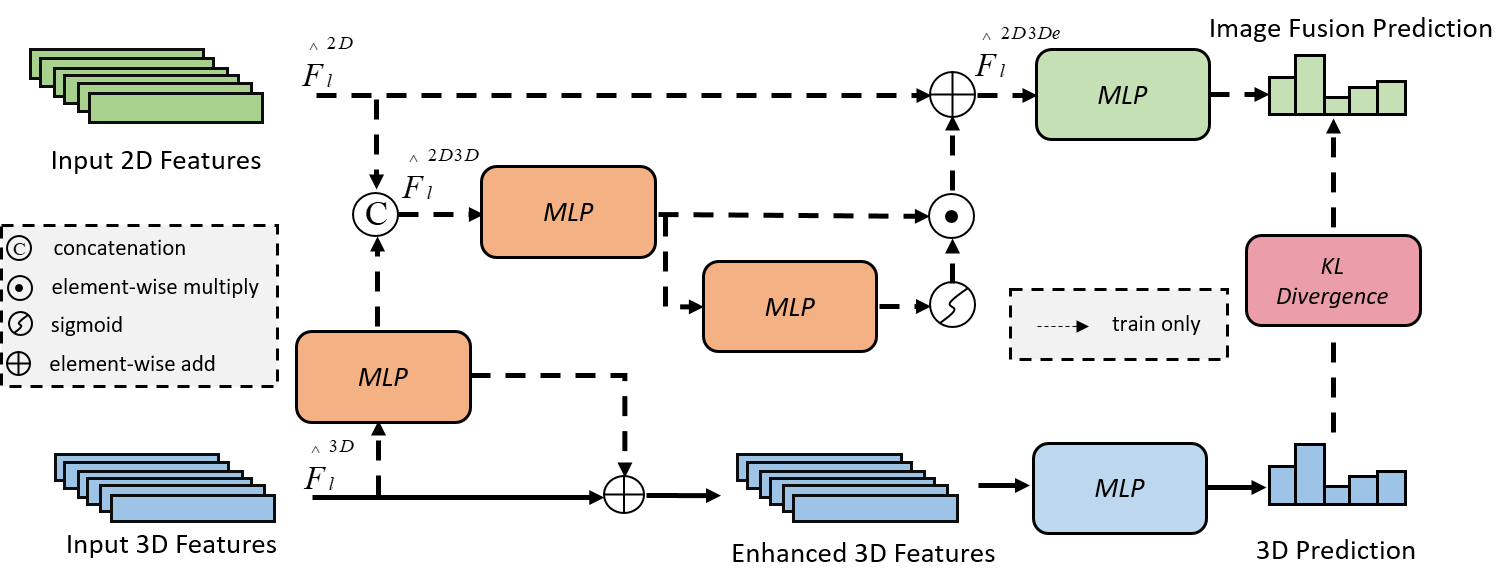}
    \captionsetup{belowskip=-10pt}
    \caption{Part of the architecture of 2DPASS. This figure shows how the 2DPASS transfer the image modal and the disposable training schema.
    During inference, only 3D branch (downside of the figure) is conducted the segmentation.}
    \label{fig:2DPASSarch}
\end{figure}

As the architecture shown in Figure \ref{fig:2DPASSarch}, for a given pair of 2D image and 3D point cloud, 
2DPASS extracts 2D features $\hat{F}_{l}^{2D} \in \mathbb{R}^{N^{img} \times D_l}$ and 3D features $F_{l}^{3D} \in \mathbb{R}^{N \times D_l}$
by 2D encoder FCN\cite{long2015fully} and 3D encoder SPVCNN\cite{tang2020searching}, since they are both pyramid networks,
the $l$ represents the feature extracted from l-th layer, the $D_l$ is the number of channels of features at l-th layer.

Because we would like to focus on the points in the image FOV,
we first select the points in the image FOV to have $\hat{F}_{l}^{3D}  \in \mathbb{R}^{N^{img} \times D_l}$ in
the 3D features $F_{l}^{3D}$,
each $\hat{F}_{l}^{3D}$ could find a corresponding 2D feature by Coordinate Transformation and Ras-
terization discussed in section 2.1.

$\hat{F}_{l}^{3D}$ are then input into a MLP layer, called \textbf{2D Learner}, the purpose of 2D Learner
is to narrow the gap between 2D and 3D features, after this, they are concatenated as fused features, written as:

\begin{equation}
    \hat{F}_{l}^{2D3D} = concat(\hat{F}_{l}^{2D}, 2DLearner(\hat{F}_{l}^{3D}))
\end{equation}

The above process shows the details of the 2D and 3D fusion in 2DPASS, since $\hat{F}_{l}^{2D}$ might not be the right pixel features corresponding to $\hat{F}_{l}^{3D}$
because of misalignment, we would like to introduce a noval way to solve this problem.

\subsection*{Deformable Fusion}
As the inspiration of \cite{xia2022vision}, we consider Deformable Attation is a powerful tool to deal with the misalignment problem,
it's designed to select the position of key and value pairs in self-attention in a data-dependent way,
where the offsets $(\Delta{x}, \Delta{y})$ are predicted by a learnable offset network.

In 2DPASS scenarios, we would like to use the projected points as ther reference points, 
and find the offset of the corresponding pixel features with image features, the formula is given as:
\begin{equation}
    (\Delta{x}, \Delta{y}) = \theta_{offset}(F_{l}^{2D}),
\end{equation}

where we follow the design of the offset network $\theta_{offset}$ in \cite{xia2022vision}, $F_{l}^{2D} \in \mathbb{R}^{H \times W \times D_l}$ is the 2D image features and $\hat{F}_{l}^{2D} \subseteq F_{l}^{2D}$,
and we could have the adaptive 2D features by:

\begin{equation}
    \tilde{F}_{l}^{2D} = \phi(F_{l}^{2D};(u + \Delta{x}, v + \Delta{y})),
\end{equation}

where $\phi$ is the bilinear interpolation function, $(u, v)$ is the coordinates of the projected pixel positions,
and the original misaligned $\hat{F}_{l}^{2D}$ is replaced by $\tilde{F}_{l}^{2D}$.

\section{Future Work}
In this paper, we address the misalignment problem in fusion task,
and how the SOTA of fusion models 2DPASS is affected by this problem, also,
we provide a possible solution to solve this problem.
However, this idea is not fully verified by experiments yet, we will leave it and the analysis of the results to the future work.

{\small
\bibliographystyle{ieeetr}
\bibliography{multimodal_challenge}
}

\end{document}